\documentclass{article}

\usepackage{arxiv}

\usepackage[utf8]{inputenc} 
\usepackage[T1]{fontenc}    
\usepackage{hyperref}       
\usepackage{url}            
\usepackage{booktabs}       
\usepackage{amsfonts}       
\usepackage{nicefrac}       
\usepackage{microtype}      
\usepackage{cleveref}       
\usepackage{lipsum}         
\usepackage{graphicx}
\usepackage{natbib}
\usepackage{doi}
\usepackage{amssymb}

\title{Facial Emotion Characterization and Detection using Fourier Transform and Machine Learning}

\author{Aishwarya~Gouru\\
	Department of Computer Science\\
	University of North Carolina at Greensboro\\
	Greensboro, NC 27402 \\
	\texttt{a\_gouru@uncg.edu } \\
	\And
	\href{https://orcid.org/0000-0003-3235-9870}{\includegraphics[scale=0.06]{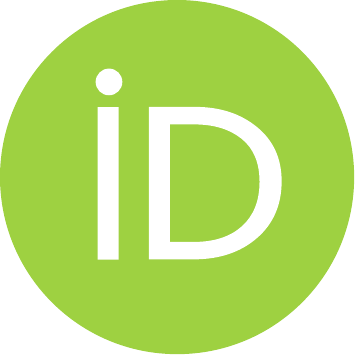}\hspace{1mm}Shan~Suthaharan\thanks{Corresponding author}} \\
        Department of Computer Science\\
	University of North Carolina at Greensboro\\
	Greensboro, NC 27402 \\
	\texttt{s\_suthah@uncg.edu} \\
}

\renewcommand{\shorttitle}{Facial Emotion Characterization and Detection}

\hypersetup{
pdftitle={A template for the arxiv style},
pdfsubject={q-bio.NC, q-bio.QM},
pdfauthor={David S.~Hippocampus, Elias D.~Striatum},
pdfkeywords={First keyword, Second keyword, More},
}

\begin{document}
\maketitle

\begin{abstract}
We present a Fourier-based machine learning technique that characterizes and detects facial emotions. The main challenging task in the development of machine learning (ML) models for classifying facial emotions is the detection of accurate emotional features from a set of training samples, and the generation of feature vectors for constructing a meaningful feature space and building ML models. In this paper, we hypothesis that the emotional features are hidden in the frequency domain; hence, they can be captured by leveraging the frequency domain and masking techniques. We also make use of the conjecture that a facial emotions are convoluted with the normal facial features and the other emotional features; however, they carry linearly separable spatial frequencies (we call computational emotional frequencies). Hence, we propose a technique by leveraging fast Fourier transform (FFT) and rectangular narrow-band frequency kernels, and the widely used Yale-Faces image dataset. We test the hypothesis using the performance scores of the random forest (RF) and the artificial neural network (ANN) classifiers as the measures to validate the effectiveness of the captured emotional frequencies. Our finding is that the computational emotional frequencies discovered by the proposed approach provides meaningful emotional features that help RF and ANN achieve a high precision scores above 93\%, on average.
\end{abstract}


\section{Introduction}
Facial expressions—such as the happy, sad, and sleepy emotions—are some of the brain's responses to psychological events. Hence, the development of computational and machine learning techniques to characterize and detect facial emotions would be useful for addressing the recently reported psychological problems and mental disorders in \cite{suthaharan2021paranoia}. In computer science discipline—especially in the last two decades—many face recognition and emotion detection techniques have been proposed \cite{ioannou2005emotion,garcia2017emotion,zeng2006spontaneous}. However, they still suffer from two major drawbacks. The first drawback is the data paucity problem. It means that it is difficult to acquire sufficient data samples for all the possible emotions such that a machine learning classifier can be trained accurately and built. The second drawback is the detection and extraction of emotional feature vectors so that a meaningful feature space can be constructed to develop machine learning models. Therefore, there is an urgent need to study facial emotion detection problems with novel approaches. Hence, this paper introduces a new terminology— the computational emotional frequency—in the Fourier domain and study its contribution to characterize emotional features.

Emotional frequency has been defined and widely used in psychology discipline \cite{higgins1997emotional}; however, the computational emotional frequency never been introduced or studied using Fourier transform for facial emotion detection. Hence, to the best of our knowledge, the proposed work brings novelty to the facial emotion recognition research. In simple terms, the computational emotional frequency may be described by the modeling and simulation of emotional frequencies. With the computational emotional frequencies, we have introduced a novel approach to uniquely characterize different emotions with distinct frequency bands by leveraging the discrete Fourier transform \cite{beaudoin2002accurate}. We validated this approach using the classification performance scores of the random forest and artificial neural network classifiers as the qualitative measures.

\section{Objectives}
\label{sect:objectives}
\begin{figure}[tb]
	\begin{centering}
	\includegraphics[width=0.67\textwidth]{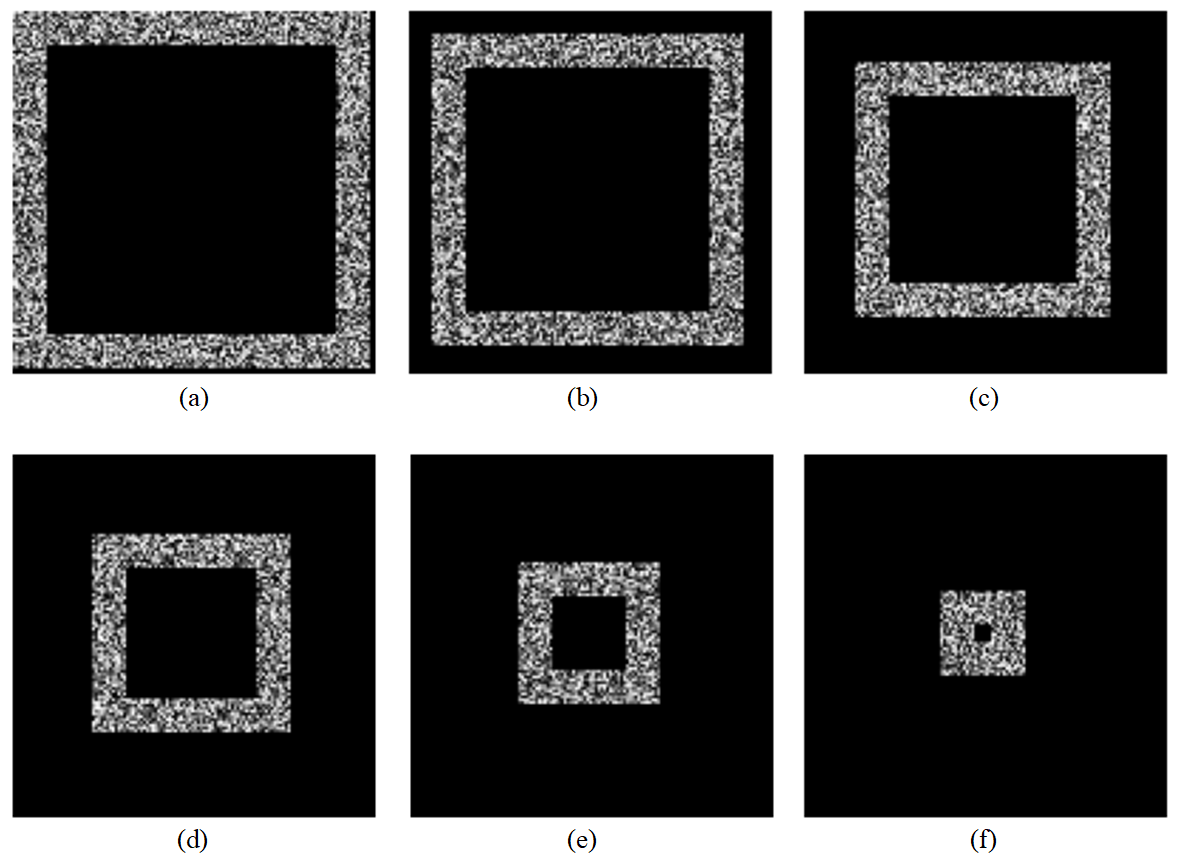}
	\caption{It illustrates some of the narrow-band frequency kernels}
	\label{fig:narrow-band}
	\end{centering}
\end{figure}

We present a Fourier-based machine learning technique that translates the emotional signals—emitted by the facial emotions—into linearized emotional frequencies in vertical and horizontal directions to characterize emotions, construct feature vectors, and detect facial emotions. The first objective is to find the origins of a range of vertical frequencies of the signal emitted by the emotional features, given a narrow-band horizontal frequencies and the same range of horizontal frequencies of the signal, given a narrow-band vertical frequencies. It includes the discovery of these emotional frequencies using the rectangular narrow-band kernels. A subset of rectangular narrow-band kernels applied in the Fourier domain are shown in Figure \ref{fig:narrow-band}. Our definition of the computational emotional frequencies is defined by spatial origins of the hidden narrow-band frequencies in the Fourier domain, which are masked by these kernels. The second objective is to construct feature vectors (or a feature space) using these analogies and study their effectiveness using performance scores of the RF and ANN classifiers, as the qualitative measures, by applying them on the feature space of the emotional features.

\section{Background}
\label{sect:background}
The current facial emotion recognition systems are evolved from the fundamental system—the Facial Action Coding System (FACS)—that helps the grouping of the facial changes based on their descriptions on the surface of the facial regions \cite{ekman1978facial}. The FACS has been later extended by Tian, Kande, and Cohn \cite{tian2001recognizing} with the concept of Action Units, which allow the integration of facial landmarks (e.g. lips, eyes, and mouth) into the system to improve its performance in terms of the grouping of the emotions (e.g., happy, anger, and sad). The integration of the landmarks creates a significant amount of uncertainties while increasing computational complexity. Hence, Viola and Jones \cite{viola2004robust} proposed an algorithm to increase the speed of computation by adapting the AdaBoost learning and cascade approach to speed up the computation along with a new concept called Integral Image that also allows faster computation in a real-time environment. In our approach, we tried to avoid the use of landmarks; however, the frequency domain techniques indirectly include the emotional frequencies emitted in the facial landmark regions. 

The random forest classifiers have been extensively studied to understand their suitability for face detection \cite{kremic2016performance,o2013facial,see2017investigation}. For example, in \cite{kremic2016performance}, authors have conducted an experimental research and studied the performance of random forest and support vector machine (SVM) in terms of detecting faces to make them suitable for mobile applications. In \cite{o2013facial}, authors mainly focused on the facial textures; hence, they combined the local patterns and random forest learning to develop models for facial recognition. In \cite{see2017investigation}, authors have implemented a technique by combining Gabor Filter and Oriented Gabor Phase Congruency Image with random forest learning. All these approaches resulted in significantly high accuracy for the random forest classification. This is one of the reasons for us to use the performance scores of the random forest as the measure to validate the proposed approach to discover emotional frequencies.

Teo, De Silva and Vadakkepat \cite{teo2004facial} developed a model for facial expression detection that uses integral projection, statistical computation, a neural network and Kalman filtering. This approach also achieves a very good accuracy. Another emotion recognition recognition technique is proposed in \cite{ioannou2005emotion} that performs facial expression analysis using the concept of neurofuzzy network that integrates psychological findings in the analysis. Authors of \cite{zeng2006spontaneous} used kernel whitening, support vector data descriptors, and Gaussian based classifiers in their proposed approach. This approach concluded that one-class classification methods can reach a good balance between the labeling and computation overheads,  and the recognition performance. Facial expression recognition problem was also studied by Otsuka and Ohya \cite{otsuka1997recognizing}, but they used the Hidden Markov Models (HMM) in a dynamic environment with multiple subjects. Authors first used a two-dimensional Fourier transformation to construct Feature vectors, by image processing techniques, and then applied HMM to represent distinct facial expressions. 

Similarly, Cohen et. al. studied the facial expression recognition problem \cite{cohen2003facial}, but they focused on video sequences; hence, they developed temporal and static models.  They used the multi-level HMM classifiers to recognize facial expression sequences and segment video sequences. In contrast, De Silva, Miyasato, and Nakatsu \cite{de1997facial} studied the facial emotion recognition problem using multimodal information of the emotions acquired from the audio, video, and hybrid clips. They assigned weights to both audio and video inputs and generated outputs based on the input information. Similarly, A technological review by Garcia-Garcia, Penichet and Lozano \cite{garcia2017emotion} discussed some of the different types of expression analysis models that are available in this research domain. This literature survey suggests none of these techniques addresses the problem of characterizing the emotions, before developing machine learning, without using the landmarks (region of interest) and extracting emotional features at various levels of narrow-band frequencies. Our goal is to fill this gap by presenting a novel approach.

\section{Methods}
\label{sect:methods}

Facial images are generally formed by the convoluted facial and emotional features, and an additive noise. Hence, we can mathematically model this example as follows:
\begin{equation}
s = u \circledast v + \epsilon 
\end{equation}

where $u$ represents a set of convoluted facial features, $v$ represents a set of convoluted emotional features, and the operator $\circledast$ represents the convolution operator. For the purpose of emotion detection, we ignore the noise term, since efficient noise filtering may be applied to remove noise. Hence, the noise-free model for facial emotion detection is:
\begin{equation}
s = u \circledast v
\label{eq:suv}
\end{equation}

To simplify the explanation, we consider two subjects that satisfy the above noise-free model.
\begin{equation}
\verb|Subject 1|: \hspace{1cm} s_1 = u_1 \circledast v_1; \hspace{0.25cm} s'_1 = u_1 \circledast v'_1,
\end{equation}
where $u_1$ represents the first subject's facial features, and $v_1$ and $v'_1$ represent two emotional features of the first subject.
\begin{equation}
\verb|Subject 2|: \hspace{1cm} s_2 = u_2 \circledast v_1; \hspace{0.25cm} s'_2 = u_2 \circledast v'_1
\end{equation}
where $u_2$ represents the second subject's facial features, and $v_1$ and $v'_1$ represent the same two emotional features of the second subject. Now suppose a facial image $s \in \{s_1, s_2\}$ that satisfies the model $s = u * v$ is given, then we could raise two important questions:
\begin{itemize}
\item{Question 1: Is it possible to extract $u$ from $s$ and determine if $u = u_1$ or $u = u_2$ by using a machine learning model? In other words, if a facial image is given, is it possible to find who is the person? This is facial recognition;}
\item{Question 2: Is it possible to extract $v$ from $s$ and determine if $v = v_1$ or $v = v'_1$ by using a machine learning model? In other words, if a facial image is given, is it possible to find what is the emotion? This is emotion detection.}
\end{itemize}
The goal of this paper is to address the challenges associated with the second question without using a facial recognition solution and propose a ML-based solution to detect facial emotions. In essence, we propose to develop a ML model $f$ such that
\begin{equation}
y = f_{\alpha, \beta}(s)
\end{equation}
where the parameter set $\alpha$ is the training parameter, $\beta$ is the hyper-parameter, and $y$ is the response variable which represents the emotion labels (e.g., sleepy, happy, or sad).

\subsection{Generalized Model}
In our proposed approach, we are mainly interested in grouping image pixels of $s$ with respect to their frequency bands (i.e., computational emotional frequencies) with the hope of revealing the emotional features that are usually convoluted in the spatial domain $s$, and latent in the frequency spectrum $S$ of $s$. The revelation of emotional features helps us construct a meaningful feature vectors and a feature space to develop machine learning models to classify facial emotions. Hence, we use FFT to transform the image signal $s$ in equation(\ref{eq:suv}) as follows:

\begin{equation}
S = T(u \circledast v),
\label{eq:sfe}
\end{equation}

where $s = u \circledast v$, $u$ is the convoluted facial features, $v$ is the convoluted emotional features,  and $T$ represents FFT. Facial landmarks generally play major roles in characterizing and detecting facial emotions in many applications. Our proposed approach does not utilize the landmarks; however, the emotional frequencies of the facial landmarks are automatically included in the frequency spectrum. Using the convolution theorem, we could write the convolutional model in equation(\ref{eq:sfe}) by the following product equation:

\begin{equation}
S = T(u) \times T(v)
\end{equation}

Hence, by applying the frequency masking technique, we could expand the above model to a consolidated parametric model for multiple subjects and emotions as follows:

\begin{equation}
F_i = M_i(S ) = M_i(T(u)) + M_i(T(v)),
\label{eq:Mi}
\end{equation}

where $M_i$ is the $i^{th}$ narrow-band frequency kernel that is parametrized and linearized with respect to the fixed width, $b$, of the narrow-band kernels and the number of kernels, $p$, $F_i$ is considered the $i^{th}$ emotional frequency, and $i = 1, 2, \dots, p$. 

\subsection{Computational Emotional Frequencies}
The emotional frequencies $F_i$s are captured in the Fourier domain using the narrow-band frequency kernel $M_i$, while computational emotional frequencies $s_i$ are defined in the spatial domain by applying the inverse FFT (i.e., iFFT), as follows,

\begin{equation}
s_i = T^{-1}(F_i) = T^{-1}(M_i(T(u))) + T^{-1}(M_i(T(v)))
\label{eq:TMi}
\end{equation}

where $s_i$ is considered the $i^{th}$ computational emotional frequency that provides the $i^{th}$ feature for the feature space, where the feature space satisfies: 

\begin{equation}
y = f_{\alpha, \beta}(s_1, s_2, …, s_p),
\end{equation}

where $f_{\alpha, \beta}$ is the machine learning model that is parametric with $\alpha$ and $\beta$ with the goal of optimizing it to develop a model to detect emotions. In this paper, we use RF and ANN for the parametric machine learning model $f_{\alpha, \beta}$. In essence it solves the problem of developing a ML model to detect facial emotions, while the equation (\ref{eq:TMi}) characterizes the facial emotions.

We empirically show, through a set of simulations with machine learning, that FFT extracts meaningful emotional features that define separable facial and emotional frequencies. For this purpose we use the performance scores of the RF and ANN classifiers as the measures of quantifying the emotional features. The goal of this paper is to estimate the computational emotional features $s_1, s_2, \dots, s_p$ using the emotional frequencies captured in frequency spectrum.

\section{Feature Learning}

\begin{figure}[tb]
	\begin{centering}
	\includegraphics[width=0.7\textwidth]{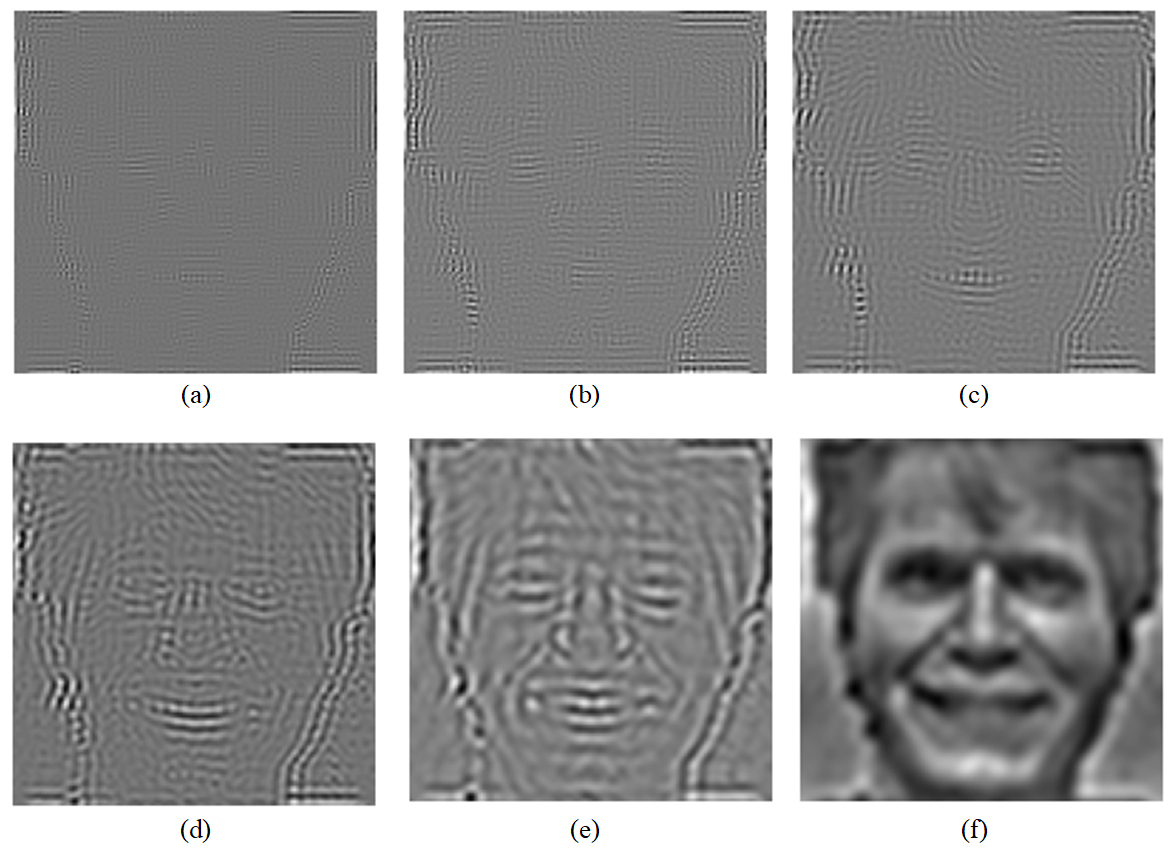}
	\caption{It illustrates some of the emotional frequencies of the happy facial expression}
	\label{fig:happy-face}
	\end{centering}
\end{figure}

We used a narrow-band frequency kernels in Frequency domain to mask out the spatial frequencies to define emotional frequencies by using equation (\ref{eq:Mi}) and detect them using equation (\ref{eq:TMi}) such that a meaningful feature space can be constructed for training and building a machine learning model. We have presented a few narrow-band frequency kernels in Figure \ref{fig:narrow-band}. This process acts as a frequency masking ($M_i$) technique that captures spatially orthogonal emotional frequencies, as described in section (\ref{sect:objectives}), with respect to given narrow-band frequencies, either vertically or horizontally. The spatial frequency features of the hidden emotions in the Fourier domain are revealed by the kernels are presented in Figures \ref{fig:happy-face} and \ref{fig:sad-face}, for the happy and sad emotions, respectively. We can clearly see the effects of these two emotional expressions, in particular, in the frequency-band presented in Figures \ref{fig:happy-face}(e) and \ref{fig:sad-face}(e).

\section{Experimental Results}
In our experimental research, we have adapted the \href{http://vision.ucsd.edu/content/yale-face-database}{Yale-Face dataset} to validate our proposed approach and build a machine learning model. We have used  the performance scores of the RF and ANN learning model as the measures to validate the computational emotional features detected by the proposed approach. As a result, we have developed an automated  machine learning model with a companion feature space to classify facial emotions. In the Yale-Face dataset, there are 9 different facial expressions with 15 subjects; hence, there are a total of 135 grayscale images. We have only used the following 5 expressions to test our hypothesis and develop a machine learning model: 1. Happy, 2. Sad, 3. Sleepy, 4. Surprised, 5. Wink. It means we have 75 images to perform our tasks and build a model. Since we don't have sufficient data to train and build a model, we have used the fast Fourier transformation and the availability of large number of frequency bands (narrow-bands) to extract many spatial frequency features to extend the number of samples with high-dimensional feature vectors.

\subsection{Feature Space Generation}
As mentioned in the previous section, we have used 75 images for the feature space construction. The first step is the application of a simple dimensionality reduction. In dimensionality reduction, we have eliminated the color channels and converted the images to grayscale. The resulting grayscale images are resized to a standard of 128 x 128 pixels, we used a Bicubic interpolation for this purpose. This step is important to generate a balanced feature space. The built-in module of the Scipy FFT library is used for the generation of Fourier transformation of the images. The algorithm presented in this paper uses FFT and iFFT. FFT converts the spatial domain (Input image) to the frequency domain. That means the input image is converted to a spectral image in the Fourier domain. Twenty five narrow-band frequency kernels are applied to the frequency domain as masks and the resulted amplified spectrum are converted back to the spatial domain using iFFT. These images display computational emotional features; hence, they are used to construct a feature space with dimension 25. When an image data is being written to feature space, each expression is labeled with a specific number.  Labeled classes are as follows, 1- Happy, 2- Sad, 3- Sleepy, 4-Surprised, 5- Wink, which means, we have built a supervised machine learning model. These labels are manually assigned to each expression. 

\subsection{Model Training}
\begin{figure}[tb]
	\begin{centering}
	\includegraphics[width=0.7\textwidth]{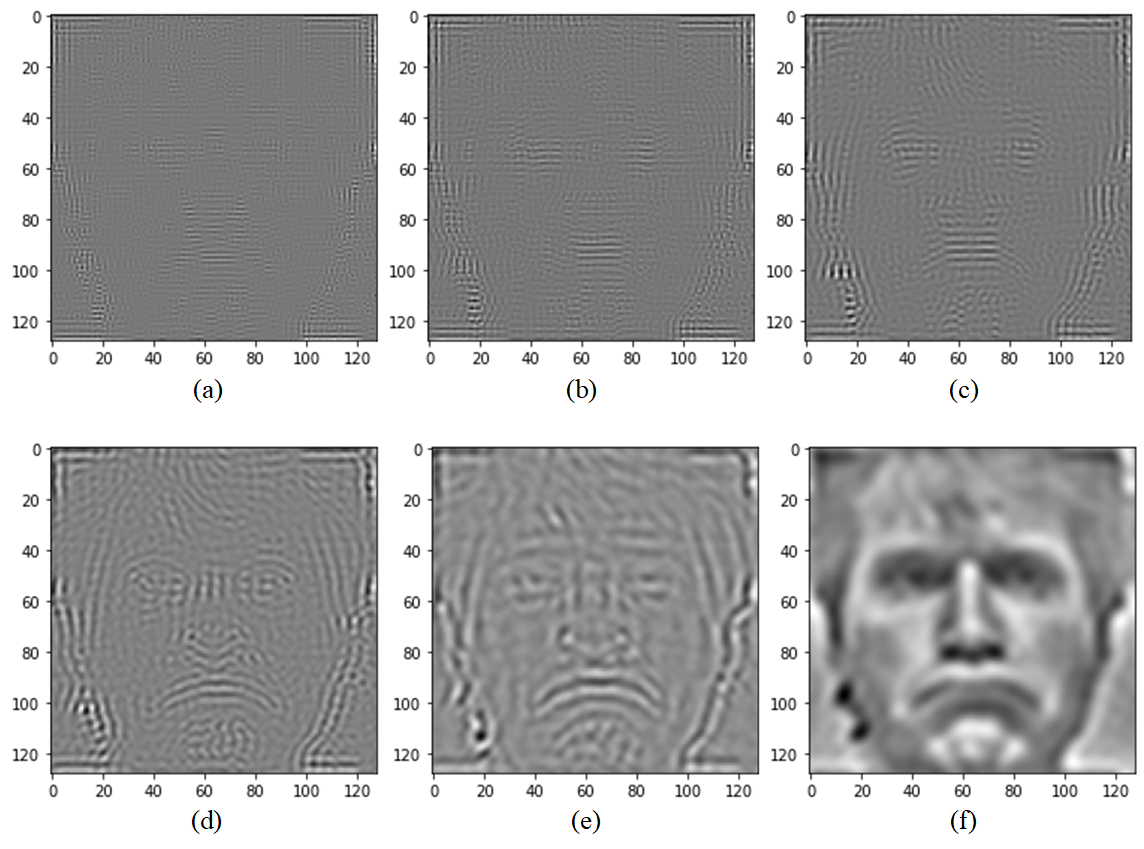}
	\caption{It illustrates some of the emotional frequencies of the sad facial expression}
	\label{fig:sad-face}
	\end{centering}
\end{figure}

For the model training, a training and testing set is built and these are used to validate the machine learning model. As mentioned earlier, the feature space generated is based on pixels, and all the image dimensions are 128 × 128; hence, the number of observations for one image is 16,384 vectors. Each image is of size 128 × 128 pixels and we used 25 frequency frames. The dimensions, 25 are achieved by choosing the alternate frames ranging from 14 to 64 that are generated by the narrow-band frequency kernels in the Fourier domain. This implies, for each image 128 × 128 × 25 = 409600 feature vectors are generated. There are 15 subjects and we selected 5 emotions; hence, we have 75 image in total. Therefore, our data domain consists of 30,720,000 feature vectors in total. Some of the frames used to generate the feature space are shown in Figures \ref{fig:happy-face} and \ref{fig:sad-face}. To train the RF model with such a huge data domain, it took an average of 30 minutes in an i5 processor when 100 trees are used. Similarly, it took about 5 hours for ANN to be trained with 75 epochs. This computational constraint is the main limitation of the proposed approach while its goal is to alleviate the data paucity problem.

\subsection{Domain Splitting}
After generating the feature space, we have implemented the RF and ANN classifiers. For the preparation of the algorithms, we have used domain splitting where a percentage of the data domain is randomized and stored in two data frames, namely, training and testing dataset. These data frames are used to implement he RF and ANN classifiers. The primary step to implementing the algorithm is to split the data into training and testing sets. In this experiment, we have used an 80:20 ratio to split the data domain. 80\% of the data domain is randomized and is used for the training of the model. The remaining 20\% of the randomized data domain is used for validating the data. The randomization helps the data to evenly populate across the data frame and help develop a better training strategy.

\subsection{Model Validation}
As we have implemented RF and ANN classifiers on the training and test set, the results obtained are based on the qualitative measures of the model. Since the data domain is a balanced dataset, we can include the accuracy measure if the validation process. We have evaluated the model based on the well-known qualitative measures, these measures are calculated based on the true positive, true negative, false positive, and false negative values. These values are used to generate accuracy, precision, sensitivity, and specificity scores \cite{suthaharan2016supervised}: 

\begin{equation}
\verb|Accuracy| = \frac{1}{1 + \frac{FP+FN}{TP+TN}} \hspace{1cm}
\verb|Precision| = \frac{1}{1 + \frac{FP}{TP}}
\end{equation}

\begin{equation}
\verb|Sensitivity| = \frac{1}{1 + \frac{FP}{TN}} \hspace{1cm}
\verb|Specificity| = \frac{1}{1 + \frac{FN}{TP}}
\end{equation}

\vspace{0.3cm}

Since the feature space is approximately balanced, we included the accuracy measure, along with other measures, for the model validation, and presented the results in Tables \ref{Tab:rf} and \ref{Tab:ann}. 

\begin{table*}[!h]
\caption{Performance scores of Random Forest model\label{tab1}}
\vspace*{6pt}
\begin{center}
\begin{tabular}{ l c c c c }
  \hline	
  Emotions & Accuracy & Precision & Specificity & Sensitivity \\
  \hline			
  Happy & 95.80 & 85.49 & 96.07 & 94.69\\
  Sad & 98.26 & 95.45 & 98.90 & 95.64\\
  Sleepy & 95.76	 & 94.64 & 85.31 & 98.67\\
  Surprised & 97.83 & 95.17 & 98.82 & 93.81\\
  Wink & 98.75 &94.92 & 98.75 & 96.31\\
  \hline  
\end{tabular}
\label{Tab:rf}
\end{center}
\end{table*}

\begin{table*}[!h]
\caption{Performance scores of Artificial Neural Network model\label{tab1}}
\vspace*{6pt}
\begin{center}
\begin{tabular}{ l c c c c }
  \hline	
  Emotions & Accuracy & Precision & Specificity & Sensitivity \\
  \hline			
  Happy & 97.06	& 93.81 & 98.44 & 91.72\\
  Sad & 97.45 & 94.22 & 98.55 & 93.09\\
  Sleepy & 96.65 & 90.52 & 97.64 & 92.62\\
  Surprised & 97.29 & 92.98 & 98.25 & 93.42\\
  Wink & 97.58 & 93.63 & 98.39 & 94.34\\
  \hline  
\end{tabular}
\label{Tab:ann}
\end{center}
\end{table*}

%

\subsection{Performance Analysis}
As the dataset contains 5 expressions, the validation results of each expression  using inbuilt measures are provided in Tables \ref{Tab:rf} and \ref{Tab:ann} for the RF and ANN models. Based on the obtained results, the primary expressions, all the expressions are accurately detected as measured by the qualitative measures presented above. The result obtained concludes that the computational emotional frequencies used to construct the feature space are meaningful and the ML developed by using them are highly efficient. For example, the precision value of 95.45\% for the sad emotion of the RF results indicate the TP is very high while the FP is low, and the precision value of 94.22\% for the same emotion of ANN also explains the same result. We can see the similar patterns for all the emotions and the models. However, one observable difference is the precision value of 85.49\% for the happy emotion of the RF results; however, it is not that significant, since since the TP is very high while the FP is very low in the case as well. Hence, overall, we can conclude the proposed approach provide a solution to characterized and detect facial emotions using Fourier transform and machine learning. 

\section{Conclusion}
The proposed Fourier-based machine learning technique was able to characterize and detect facial emotions efficiently, which is evidenced by the high qualitative performance measures of the RF and ANN classifiers. The proposed approach was also able to extend the data samples to tackle the high-dimensionality of the emotional features, and the limitations in the availability of data for learning. It was also able to identify the hidden emotional frequency features and extract them to linearize convoluted emotional features. However, it still suffers from the computational drawback because of the very high dimensional system that it creates. Our future research will focus on the development of low-dimensional structures from the high-dimensional system to build machine learning classifiers.

\section{Author Contributions}
\label{sect:contrib}
\textbf{Gouru} contributed to the literature survey, computer simulation of the design and the models, and the development of the results; writing/revising of manuscript (particularly to the background and the experimental results sections). \textbf{Suthaharan} contributed to the development of the ideas that include the emotional frequencies, their spatial correspondence through feature vectors, and their applications to machine learning; design, modeling, and development of the experimental studies; interpretation of ML results; writing/revising of manuscript.


\bibliographystyle{unsrtnat}
\bibliography{references}  

\begin{thebibliography}{17}
\providecommand{\natexlab}[1]{#1}
\providecommand{\url}[1]{\texttt{#1}}
\expandafter\ifx\csname urlstyle\endcsname\relax
  \providecommand{\doi}[1]{doi: #1}\else
  \providecommand{\doi}{doi: \begingroup \urlstyle{rm}\Url}\fi

\bibitem[Suthaharan et~al.(2021)Suthaharan, Reed, Leptourgos, Kenney,
  Uddenberg, Mathys, Litman, Robinson, Moss, Taylor,
  et~al.]{suthaharan2021paranoia}
Praveen Suthaharan, Erin~J Reed, Pantelis Leptourgos, Joshua~G Kenney, Stefan
  Uddenberg, Christoph~D Mathys, Leib Litman, Jonathan Robinson, Aaron~J Moss,
  Jane~R Taylor, et~al.
\newblock Paranoia and belief updating during the covid-19 crisis.
\newblock \emph{Nature Human Behaviour}, 5\penalty0 (9):\penalty0 1190--1202,
  2021.

\bibitem[Ioannou et~al.(2005)Ioannou, Raouzaiou, Tzouvaras, Mailis, Karpouzis,
  and Kollias]{ioannou2005emotion}
Spiros~V Ioannou, Amaryllis~T Raouzaiou, Vasilis~A Tzouvaras, Theofilos~P
  Mailis, Kostas~C Karpouzis, and Stefanos~D Kollias.
\newblock Emotion recognition through facial expression analysis based on a
  neurofuzzy network.
\newblock \emph{Neural Networks}, 18\penalty0 (4):\penalty0 423--435, 2005.

\bibitem[Garcia-Garcia et~al.(2017)Garcia-Garcia, Penichet, and
  Lozano]{garcia2017emotion}
Jose~Maria Garcia-Garcia, Victor~MR Penichet, and Maria~D Lozano.
\newblock Emotion detection: a technology review.
\newblock In \emph{Proceedings of the XVIII international conference on human
  computer interaction}, pages 1--8, 2017.

\bibitem[Zeng et~al.(2006)Zeng, Fu, Roisman, Wen, Hu, and
  Huang]{zeng2006spontaneous}
Zhihong Zeng, Yun Fu, Glenn~I Roisman, Zhen Wen, Yuxiao Hu, and Thomas~S Huang.
\newblock Spontaneous emotional facial expression detection.
\newblock \emph{J. Multim.}, 1\penalty0 (5):\penalty0 1--8, 2006.

\bibitem[Higgins et~al.(1997)Higgins, Shah, and Friedman]{higgins1997emotional}
E~Tory Higgins, James Shah, and Ronald Friedman.
\newblock Emotional responses to goal attainment: strength of regulatory focus
  as moderator.
\newblock \emph{Journal of personality and social psychology}, 72\penalty0
  (3):\penalty0 515, 1997.

\bibitem[Beaudoin and Beauchemin(2002)]{beaudoin2002accurate}
Normand Beaudoin and Steven~S Beauchemin.
\newblock An accurate discrete fourier transform for image processing.
\newblock In \emph{Object recognition supported by user interaction for service
  robots}, volume~3, pages 935--939. IEEE, 2002.

\bibitem[Ekman and Friesen(1978)]{ekman1978facial}
Paul Ekman and Wallace~V Friesen.
\newblock Facial action coding system consulting psychologists press.
\newblock \emph{Palo Alto, CA}, 1978.

\bibitem[Tian et~al.(2001)Tian, Kanade, and Cohn]{tian2001recognizing}
Y-I Tian, Takeo Kanade, and Jeffrey~F Cohn.
\newblock Recognizing action units for facial expression analysis.
\newblock \emph{IEEE Transactions on pattern analysis and machine
  intelligence}, 23\penalty0 (2):\penalty0 97--115, 2001.

\bibitem[Viola and Jones(2004)]{viola2004robust}
Paul Viola and Michael~J Jones.
\newblock Robust real-time face detection.
\newblock \emph{International journal of computer vision}, 57\penalty0
  (2):\penalty0 137--154, 2004.

\bibitem[Kremic and Subasi(2016)]{kremic2016performance}
Emir Kremic and Abdulhamit Subasi.
\newblock Performance of random forest and svm in face recognition.
\newblock \emph{Int. Arab J. Inf. Technol.}, 13\penalty0 (2):\penalty0
  287--293, 2016.

\bibitem[O'Connor and Roy(2013)]{o2013facial}
Brian O'Connor and Kaushik Roy.
\newblock Facial recognition using modified local binary pattern and random
  forest.
\newblock \emph{International Journal of Artificial Intelligence \&
  Applications}, 4\penalty0 (6):\penalty0 25, 2013.

\bibitem[See et~al.(2017)See, Noor, Low, and Liew]{see2017investigation}
YC~See, NM~Noor, JL~Low, and Eugene Liew.
\newblock Investigation of face recognition using gabor filter with random
  forest as learning framework.
\newblock In \emph{TENCON 2017-2017 IEEE Region 10 Conference}, pages
  1153--1158. IEEE, 2017.

\bibitem[Teo et~al.(2004)Teo, De~Silva, and Vadakkepat]{teo2004facial}
WK~Teo, Liyanage~C De~Silva, and Prahlad Vadakkepat.
\newblock Facial expression detection and recognition system.
\newblock \emph{Journal of The Institution of Engineers, Singapore},
  44\penalty0 (3), 2004.

\bibitem[Otsuka and Ohya(1997)]{otsuka1997recognizing}
Takahiro Otsuka and Jun Ohya.
\newblock Recognizing multiple persons' facial expressions using hmm based on
  automatic extraction of significant frames from image sequences.
\newblock In \emph{Proceedings of International Conference on Image
  Processing}, volume~2, pages 546--549. IEEE, 1997.

\bibitem[Cohen et~al.(2003)Cohen, Sebe, Garg, Chen, and Huang]{cohen2003facial}
Ira Cohen, Nicu Sebe, Ashutosh Garg, Lawrence~S Chen, and Thomas~S Huang.
\newblock Facial expression recognition from video sequences: temporal and
  static modeling.
\newblock \emph{Computer Vision and image understanding}, 91\penalty0
  (1-2):\penalty0 160--187, 2003.

\bibitem[De~Silva et~al.(1997)De~Silva, Miyasato, and Nakatsu]{de1997facial}
Liyanage~C De~Silva, Tsutomu Miyasato, and Ryohei Nakatsu.
\newblock Facial emotion recognition using multi-modal information.
\newblock In \emph{Proceedings of ICICS, 1997 International Conference on
  Information, Communications and Signal Processing. Theme: Trends in
  Information Systems Engineering and Wireless Multimedia Communications
  (Cat.}, volume~1, pages 397--401. IEEE, 1997.

\bibitem[Suthaharan(2016)]{suthaharan2016supervised}
Shan Suthaharan.
\newblock Supervised learning algorithms.
\newblock In \emph{Machine learning models and algorithms for big data
  classification}, pages 183--206. Springer, 2016.

\end{thebibliography}






\end{document}